\documentclass[english, letterpaper, 10pt]{article}

\usepackage[T1]{fontenc}
\usepackage{textcomp}

\usepackage{graphicx}
\usepackage{amssymb,amsmath}
\usepackage{fixltx2e}
\usepackage{fancyhdr}
\usepackage{url}

\usepackage[numbers,sort&compress]{natbib}

\usepackage{dsfont}
\usepackage[top=1in, bottom=1.25in, left=0.875in, right=0.875in]{geometry}
\usepackage{bm}

\usepackage{mathptmx}   
\usepackage{titlesec}
\usepackage{titling}

\usepackage{booktabs}
\usepackage{graphicx}
\usepackage{multirow}
\usepackage{subcaption}

\usepackage[english]{babel}
\usepackage[utf8]{inputenc}
\usepackage{amsmath}
\usepackage{amsfonts}
\usepackage{graphicx}
\usepackage{enumitem}
\usepackage[colorinlistoftodos]{todonotes}
\usepackage{abstract}

\usepackage{indentfirst}

\usepackage{times}

\usepackage{titlesec}
\titleformat{\section}{\bfseries\centering\fontsize{11pt}{13pt}\selectfont}{\thesection.}{4pt}{\uppercase}
\titleformat{\subsection}{\bfseries\fontsize{11pt}{13pt}\selectfont}{\thesubsection}{6pt}{}
\titlespacing*{\section}{0pt}{10pt plus 8pt}{4pt}
\titlespacing*{\subsection}{0pt}{6pt}{3pt}

\usepackage{titling}
\pretitle{\vspace{-15mm}\begin{center}\bfseries\LARGE}
\posttitle{\end{center}}
\preauthor{\begin{center}\large}
\postauthor{\end{center}\vspace{-15mm}}

\usepackage{etoolbox}
\apptocmd{\thebibliography}{\setlength{\itemsep}{-2pt}}{}{}

\newlength{\myitemsep}
\setlist[itemize]{itemsep=-1pt, topsep=2pt}
\setlist[enumerate]{itemsep=-1pt, topsep=1pt}

\title{Automatic Construction of a Recurrent Neural Network based Classifier for Vehicle Passage Detection}

\author{Evgeny Burnaev$^{1,2}$, Ivan Koptelov$^{2}$, German Novikov$^{2}$, Timur Khanipov$^{2}$
\\
$^1$Skoltech, $^2$Institute for Information Transmission Problems RAS}

\date{}

\begin{document}
\pagenumbering{gobble}
\maketitle

\renewcommand{\abstractname}{\vspace{0pt}\fontsize{11pt}{13pt}\selectfont \uppercase{Abstract}\vspace{-4pt}}
\begin{abstract}
\normalsize
Recurrent Neural Networks (RNNs) are extensively used for time-series modeling and prediction. We propose an approach for automatic construction of a binary classifier  based on  Long Short-Term Memory RNNs (LSTM-RNNs) for detection of a vehicle passage through a checkpoint. As an input to the classifier we use multidimensional signals of various sensors that are installed on the checkpoint. Obtained results demonstrate that the previous approach to handcrafting  a classifier, consisting of a set of deterministic rules, can be successfully replaced by an automatic RNN training on an appropriately labelled data.
\vspace{6pt}
\end{abstract}

\section{Introduction}
\label{sec:introduction}

Paper \cite{Koptelov14} describes an Automatic Vehicle Classifier (AVC) for toll roads, based on video classification and installed on most of Russian toll roads. Vehicle Passage Detector (VPD) is one of the most important parts of the AVC system. VPD uses as input several binary signals from other AVC subsystems (binary detectors), and makes decisions about a vehicle passage (whether a vehicle is located in the checkpoint area). VPD is based on a ``voting scheme'': a vehicle passage is detected if most of the binary detectors provide a positive answer. This logic is augmented by a set of empirical rules, provided by a human expert to quantify and take into account time delays between switches of the binary signals, properties of a sequence of these switches and other information. These rules were extended and modified during AVC test deployment based on an analysis of encountered errors.

The previous paper, devoted to VPD in AVC \cite{bocharov15}, states that the VPD accuracy is $99.58\%$. Since then the test dataset has been extended by new detection and classification error cases. It should be noted that in the current paper we use tests which run with a disabled trailer coupler detector. AVC version, described in the previous paper, provides $83.47\%$ accuracy on the new dataset if the coupler detector is disabled. At the same time the current classifier version provides $88.90\%$ accuracy without the coupler detector and $91.10\%$ with it. Comparing to the previous version the new classifier has optimized algorithms for a shield and a trailer couplers detection, a correlational detector and an updated fusion method for binary detectors aggregation.

Creating rules of this type is a painstaking job requiring creative approach. It would be interesting to develop an automatic method where a machine learning algorithm could replace a human expert. This approach potentially could also produce a higher classification quality. Therefore, in this paper we solve the problem of creating a method for automating an AVC synthesis and minimizing a human involvement.

\section{Data description}
\label{sec:data_desc}

The input data consists of AVC log records. Each file contains one or several vehicles passages. All the numerical experiments are conducted on a dataset consisting of $4761$ log files. The system log is filled with three-dimensional signal samples $X_t$, each component of which is binarized and produced by one of the following sensors: a correlational detector (sensitive to changes in video-stream images), an induction loop (mounted inside a lane and sensitive to a metal), a shield detector (detects occlusions of a shield, located opposite to a camera). Also the records contain a frame sequence number, manually created reference signal (labels) and predictions of a basic classifier, which is based on human tweaked rules \cite{Koptelov14}. A file record is created and saved in a database if and only if at least one of the input signals has changed.

\begin{table}[h]
\begin{center}
	 \begin{tabular}{|l|r|r|r|r|r|}
	 \hline
	 	Frame  &  Shield &  Loop &  Cor &  Basic clf & Ref. pass \\
	 	No. & & & & & \\
		\hline
		196 &       1 &     0 &    0 &     0 &  0 \\
		201 &       1 &     1 &    0 &     0 &  0 \\
		202 &       0 &     1 &    1 &     1 &  0 \\
		208 &       1 &     1 &    1 &     1 &  1 \\
		246 &       0 &     1 &    1 &     1 &  1 \\
		266 &       1 &     1 &    1 &     1 &  0 \\
		268 &       1 &     1 &    0 &     1 &  0 \\
		269 &       1 &     0 &    0 &     0 &  0 \\
		270 &       0 &     0 &    0 &     0 &  0 \\
	 \hline
	\end{tabular}
	\caption{A log and a labeled sample}
\label{tab:tab1}
\end{center}
\end{table}

\textbf{Table 1} shows a log sample. Notations:
 	\textbf{Shield} --- a shield detector, 
	\textbf{Loop} --- an induction loop,
	\textbf{Cor} --- a correlational detector,
	\textbf{Basic clf} --- a basic classifier prediction,
	\textbf{Ref. pass} --- a reference passage.

\section{Quality evaluation}
The problem specifics dictates that a special quality passage evaluation metric should be used, which is equal to the standard pointwise two-class classification metric \textbf{Accuracy} only in extreme cases. The reason is that the standard metrics using pointwise difference between a reference signal and a predicted signal are not able to estimate a passage classification quality from a physically sensible viewpoint. The metric used in this paper is a \textbf{Pass Quality (PQ)}:
$
    PQ = \frac{R}{R + \sum Err},
$
where $R$ is a number of correctly detected passages, while $\sum Err$ is a sum of classification error costs on a whole test dataset. Calculation of $\sum Err$ for various error types (missed passage, merged passages, etc.) is a complicated procedure described in \cite{Malugina15}, see also table \ref{tab:tab2}. Here $L$ denotes the true number of passages in the analysed test signal, $K$ denotes the number of detected passages (see details in \cite{Malugina15}).
\begin{table}[h]\begin{center}
	\begin{tabular}{|c|c|c|c|}
	\hline 
	Ref. pass. & Detected pass. & Err. Desc. & Err. Weight \\
	\hline 
	1 & 1 & no error & 0 \\ 
	\hline
	1 & 0 & missed passage & 1 \\ 
	\hline
	0 & 1 & false passage  & 1 \\ 
	\hline
	L & 1 & merged passages & L \\ 
	\hline
	1 & K & split passage  & K \\ 
	\hline
	L & K & multiple error & max(L, K) \\ 
	\hline
	\end{tabular} 
\caption{Classification error costs}
\label{tab:tab2}\end{center}   
\end{table}

This quality metric does not take into account how far the detected passage is shifted from the ideal one. However the conducted experiments show that this is not needed for applications, since the sequence of correct passages and their intersections with real passages at least in one instant are important, not the delays themselves.

\section{Exploratory analysis}
\label{subsec:expan}

In this section we compare results obtained with various machine learning methods: gradient tree boosting \textbf{XGB} (see description of the algorithm in \cite{Tianqi}; the implementation from \cite{xgb16} was used), logistic regression \textbf{LR} from the \textbf{scikit-learn} package \cite{sklearn11}, fully connected neural network \textbf{NN}  with a one hidden layer consisting of $12$ neurons (see approaches to training such networks in \cite{init, boosting, HDA, HDA2}), simple recurrent neural network \textbf{SimpleRNN} from the \textbf{Keras} package \cite{keras16}.

Due to an atypical task partitioning of the data into the training and the control sets were organized as follows: logs for a respective set are randomly selected in a random order, but the order of frames inside every log remains unchanged.

Results for training on the source signal $X_t$ without accounting for past values are provided in table \ref{tab:tab3}, see lines $\mathbf{XGB}_1,$ $\mathbf{LR}_1,$ $\mathbf{NN}_1$. The obtained results imply that due to the autocorrelation of $X_t$, classification using standard methods without taking into account the dependency of $X_t$ on $X_{t-1}, \dots X_{t-k}, \dots$, provides low \textbf{PQ} values. The situation that \textbf{PQ} values are comparable can be easily explained by the fact that at each moment of time the three-dimensional vector $X_t$ can take only eight distinct values, thus in the considered case all methods can easily provide similar decision rules.

The first idea to increase classification quality w.r.t. the \textbf{PQ} metric is to extend the feature space $X_t$ by using previous values $\{X_{t-1}, \ldots, X_{t-w}\}$. The results of these experiments are provided in table \ref{tab:tab3}, see lines $\mathbf{XGB}_2,$ $\mathbf{LR}_2$, $\mathbf{NN}_2$. The optimal window size $w$ for each classifier type is selected using the cross-validation procedure. The threshold, producing a binary signal from a classifier output probability (provided by the logistic regression and the neural network), is selected by maximizing the objective function on the training set. One may see that this extension provides significant increase in quality. The $\mathbf{NN}_2$ gives the best result. This result, however, is not better than that of the basic classifier.

The second natural idea is to use recurrent neural networks, which proved to be productive in labelling of sequences and time series classification. We will use the standard recurrent network \textbf{SimpleRNN} with a one recurrent layer and a one hidden layer \cite{haykin}. It turns out that this model allows to obtain a higher quality which is pretty close to that of the basic classifier.

\begin{table}[h]
 \begin{center}
	\begin{tabular}{|c|c|c|c|c|}
	\hline 
	Classifier & R & $\sum Err$ & PQ\\ 
	\hline 
	$\mathbf{XGB}_1$ & \textbf{3090.2} & 2802.3 & 0.532 \\
	$\mathbf{LR}_1$ & \textbf{3090.2} & 2802.3 & 0.532 \\ 
	$\mathbf{NN}_1$ & \textbf{3090.2} & 2802.3 & 0.532 \\ 
	$\mathbf{XGB}_2$ & 1806.0 & 453.0 &  0.799  \\ 
	$\mathbf{LR}_2$ & 1794.1 & 302.6 & 0.856  \\ 
	$\mathbf{NN}_2$ & 1758.0 & 276.6 & 0.864 \\  
	$\mathbf{SimpleRNN}$ & 1784.8 & \textbf{214.3} & \textbf{0.892} \\
	\hline
	$\mathbf{Basic}$ $\mathbf{Classifier}$ & 1684.3 & 158.7 & 0.914 \\  
	\hline
	\end{tabular}
	\caption{Comparison of classifier models}
 \label{tab:tab3}  
 \end{center}
\end{table}

\textbf{Note}: the $\mathbf{Basic}$ $\mathbf{Classifier}$ uses an additional feature from the trailer coupler detector. This feature allows to distinguish between vehicles moving at a close distance and vehicles with trailers thus increasing the basic method accuracy \textbf{PQ} from $0.889$ to $0.914$. In our experiments we do not use this feature, although this additional information could potentially increase classification quality.

The primary reason for a low prediction accuracy is that all algorithms optimize not the target quality metric but a different value --- the mean squared prediction error \textbf{MSE}, which is not in a good correspondence with the target metric \textbf{PQ}. Hereinafter by an error we mean $\mathbf{PQE} = 1 - \mathbf{PQ}$.

\section{Classification based on RNNs}

From the results of section \ref{subsec:expan} we can notice that only the RNN model provides results with the accuracy comparable to that of the baseline classifier, constructed manually by collecting and ensembling rules, distributed in time. Thus, RNNs is a promising approach to automate construction of classifiers and further improve the accuracy of VPD.

\subsection{RNN Architecture}

\begin{figure}[!ht]
  \centering
    \includegraphics[scale=0.183]{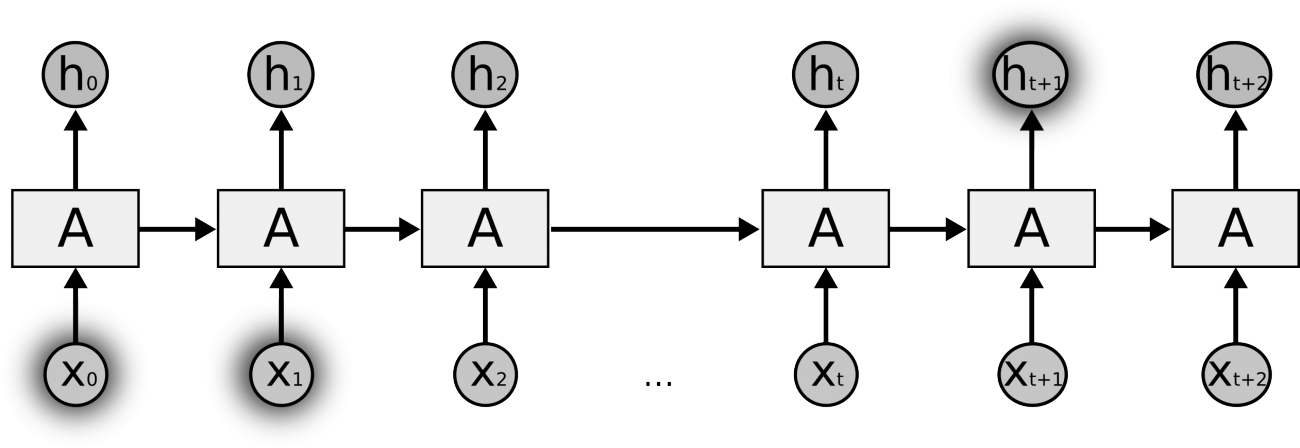}
  \caption{Long Range Dependence in RNN}
\label{pic:long_term_dep}
\end{figure}

The original recurrent neural network model \textbf{SimpleRNN} contains only one hidden layer; the output signal of each neuron is used as an input to the same neuron at the next moment of time. In case an input signal has a long duration, the exploding and the vanishing gradient problem appears when training the model and calculating gradients of a neural network performance function, see \cite{bengio94} for more details. This effect stems from the fact that the gradients of the RNN's performance function depend on the product of the gradients of neurons in the hidden layer, calculated for all successive values of the training signal, see Fig. \ref{pic:long_term_dep}; as a consequence, this product can take big absolute values as well as tends to zero.

One approach to avoid this effect is to use the Long Short-Term Memory Neural Network architecture (\textbf{LSTM}), which allows effective modelling of long range dependences in signals.

In \cite{cho14} ($2014$) the authors proposed the neural network architecture called Gated Recurrent Unit (\textbf{GRU}), based on the same principles as the \textbf{LSTM}, but it uses a more economical parametrization and fewer operations to compute an output signal. A more extensive overviews of  \textbf{RNN} architectures can be found in  \cite{rnns15, lipton15}.

\subsection{Selection of \textbf{RNN} Architecture}

To use \textbf{RNN} in practice, it is necessary to search for its optimal architecture. This section describes results on this matter. In experiments we use the following types of \textbf{RNN}: \textbf{LSTM}, \textbf{GRU} and \textbf{SimpleRNN}. All experiments are performed using the \textbf{Keras} framework \cite{keras16}, which is a wrapper of the \textbf{Python}-libraries \textbf{Theano} \cite{theano16} and \textbf{TensorFlow} \cite{tf15}.

For each of the \textbf{RNN} types we conduct a series of experiments: we play with network hyper-parameters, activation function types and a window length $w$. Since the training time of an \textbf{RNN} with several layers is rather big, we fix the number of hidden layers to be equal to one, number of neurons in each layer is limited from above by eight, the maximum number of learning epochs is limited from above by $40$. In table \ref{tab:tab_rnns} we provide results (averaged over $30$ experiments) of optimal architecture selection for each \textbf{RNN} type. One may see that selection of hyper-parameters, even in a limited space significantly improves the quality of \textbf{SimpleRNN} model, cf. with the first experiments presented in table \ref{tab:tab3}. According to these results we decide to continue to use architecture containing \textbf{LSTM} layers, since it provides the highest performance. 

\begin{table}[h]
 \begin{center}
	\begin{tabular}{|c|c|c|c|}
	\hline 
	 Model & R & $\sum Err$ & PQ\\
	\hline 
	\textbf{SimpleRNN} & 1723.0 & 161.1 & 0.915 \\
	\textbf{LSTM} & \textbf{1751.8} & \textbf{145.3} & \textbf{0.923} \\ 
	\textbf{GRU} & 1699.0 & 175.0 & 0.907 \\ 
	\hline
	\hline
	$\mathbf{Basic}$ $\mathbf{Classifier}$ & 1684.3 & 158.7 & 0.914 \\ 
	\hline
	\end{tabular} 
	\caption{Comparison of different \textbf{RNN} models}
	\label{tab:tab_rnns}
 \end{center}
\end{table}

\subsection{Further improvements}

Since when training a neural network we optimize a mean square point-wise error, which is not appropriate for the considered problem, in this section we consider various additional approaches to improve further the detection accuracy, evaluated by the \textbf{PQ} value. In particular, we consider the following tweaks: weighting of the mean square error, used as a performance function when training a neural network; smoothing the input signal by a morphological filter \cite{morph}; adding a penalty on a derivative of the neural network output, used when calculating the performance function; optimization of a threshold value, used to binarize the output signal, according to the target quality criterion \textbf{PQ} on the training set; applying the morphological filter to the neural network output signal.

\begin{figure}[!ht]
  \centering
    \includegraphics[scale=0.45]{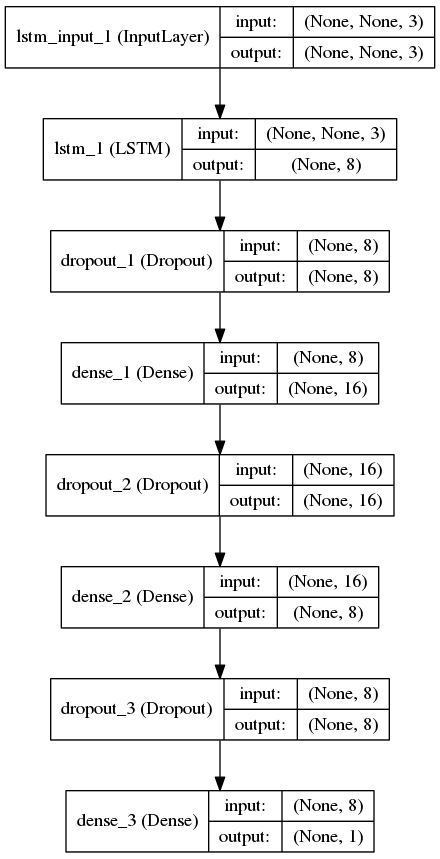}
  \caption{Final LSTM-RNN model}
\label{pic:final_model}
\end{figure}

It turned out that improvement of the detection accuracy can be obtained when expanding significantly the structure of the \textbf{LSTM} neural network. In fig. \ref{pic:final_model} we provide the modified architecture we use: one input \textbf{LSTM}-layer and two hidden \textbf{Dense} layers. However, if we do not use \textbf{Dropout} transformation, despite the fact that the standard error \textbf{MSE} decreases on the validation sample up to $0.06-0.07$, the target error \textbf{PQE} increases on the test set. In turn, if we use \textbf{Dropout} transformation before each \textbf{Dense} layer, then \textbf{MSE} increases up to $0.13$, but at the same time the target error \textbf{PQE} decreases for about $0.005-0.010$.

Also we can achieve additional significant improvement of the detection accuracy by selecting the threshold value of the output signal via the cross-validation procedure on the training set and the subsequent application of the morphological filter to the neural network output, binarized using the selected threshold value.

Also we evaluate an importance of each input signal component by estimating its influence on the accuracy of the final model. In table \ref{tab:tab4} we provide values of the performance  criterion for models, constructed using all possible combinations of input features. We can see that the best set of features is a pair $(\mathbf{Shield}, \mathbf{Cor})$.

\begin{table}[h]
 \begin{center}
	\begin{tabular}{|c|c|c|c|}
	\hline 
	 Features & R & $\sum Err$ & PQ\\
	\hline 
	$\mathbf{Cor}$ & 1714.1 & 153.9 & 0.918 \\
	$\mathbf{Loop}$ & 1621.6 & 255.9 & 0.864 \\ 
	$\mathbf{Shield}$ & 1679.5 & 200.8 & 0.893 \\ 
	$(\mathbf{Loop}, \mathbf{Cor})$ &  1713.4 & 170.6 & 0.910\\
	$(\mathbf{Shield}, \mathbf{Loop})$ &  1738.7 & 130.8 & 0.930\\
	$(\mathbf{Shield}, \mathbf{Cor})$ &  \textbf{1769.9} & \textbf{91.8}  & \textbf{0.952} \\
	$(\mathbf{Shield}, \mathbf{Loop}, \mathbf{Cor})$ &  1767.6 & 93.2  & 0.950 \\
	\hline
	\hline
	$\mathbf{Base}$ $\mathbf{Classifier}$ & 1684.3 & 158.7 & 0.914 \\ 
	\hline
	\end{tabular} 
	\caption{Accuracy of the final LSTM-RNN model for different subsets of input signal components}
	\label{tab:tab4}
 \end{center}
\end{table}

\section{Conclusions} 

We can achieve significantly better quality of classification equal to $0.952$ using only two input features, whereas in order to achieve the detection performance \textbf{PQ} equal to $0.914$, the original (handcrafted) classifier takes as input additional fourth feature from the trailer coupler detector, without which the performance drops to $0.889$. Thus, in this study we developed the automated approach for constructing and training a classifier which is superior in terms of VPD performance to the previously constructed classifiers. At this stage, further research is possible in several directions.

First, we can increase the classification accuracy by a direct optimization of the \textbf{PQ} criterion when training the neural network. The implementation of such learning algorithm is possible through the use of gradient-free optimization algorithms.

Second, we can create an aggregating mechanism for calculating a final decision from outputs of different detectors.

And finally, we can implement an integrated solution in order to eliminate the initial input data pre-processing, provided by classical image recognition methods and other
additional steps of data processing, which happen between the event ``a vehicle is shot with a camera'' and the event ``a binarized input signal $X_t$ is produced''. In other words, we propose to use the following neural network structure, which realizes all VPD subsystems by a single stack of convolutional neural networks and RNNs:
\begin{itemize}
\item On the first level we use a set of convolutional neural networks, processing images from all available cameras to extract features;
\item On the next levels features, extracted by the set of convolutional neural networks, are combined through the \textbf{RNN} architecture with signals, obtained from the induction loop and other devices of the AVC system;
\item Finally, the \textbf{RNN} type model with the structure similar to the one, shown in Fig. \ref{pic:final_model}, is used for passage detection.
\end{itemize}

\textbf{Acknowledgements:} The work of the first author was supported by the RFBR grants 16-01-00576 A and 16-29-09649 ofi\_m. The work of the other authors was conducted in IITP RAS and supported solely by the Russian Science Foundation grant (project 14-50-00150).

\bibliographystyle{plainnat}
\renewcommand\refname{\uppercase{References}}

\end{document}